\documentclass[11pt,a4paper]{article}

\usepackage[margin=2.5cm]{geometry}
\usepackage{times}
\usepackage[T1]{fontenc}
\DeclareUnicodeCharacter{2713}{\checkmark}
\usepackage{microtype}
\usepackage{amsmath,amssymb}
\usepackage{booktabs}
\usepackage{graphicx}
\usepackage{xcolor}
\usepackage{hyperref}
\usepackage{url}
\usepackage{cleveref}
\usepackage{enumitem}
\usepackage{listings}
\usepackage{tcolorbox}
\usepackage{multicol}
\usepackage{caption}
\usepackage{subcaption}
\usepackage{tikz}
\usetikzlibrary{shapes.geometric, arrows.meta, positioning, fit, backgrounds, calc}

\definecolor{apex-dark}{RGB}{26,26,46}
\definecolor{apex-red}{RGB}{233,69,96}
\definecolor{apex-l1}{RGB}{192,57,43}
\definecolor{apex-l2}{RGB}{36,113,163}
\definecolor{apex-l3}{RGB}{30,132,73}
\definecolor{apex-light}{RGB}{247,249,252}
\definecolor{apex-border}{RGB}{221,227,236}

\tcbuselibrary{skins,breakable,listings}
\newtcolorbox{apexbox}[1][]{
  enhanced, breakable,
  colback=apex-light, colframe=apex-dark,
  left=6pt, right=6pt, top=4pt, bottom=4pt,
  fonttitle=\bfseries\small, #1}
\newtcolorbox{resultbox}{
  enhanced, breakable,
  colback=green!5, colframe=apex-l3!70,
  left=6pt, right=6pt, top=4pt, bottom=4pt}
\newtcolorbox{gapbox}{
  enhanced, breakable,
  colback=blue!4, colframe=apex-l2!70,
  left=6pt, right=6pt, top=4pt, bottom=4pt}
\newtcolorbox{algorithmbox}[2][]{
  enhanced, breakable,
  colback=gray!4, colframe=apex-dark,
  title={\textbf{Algorithm #2}},
  fonttitle=\small\bfseries,
  attach boxed title to top left={yshift=-2mm, xshift=4mm},
  boxed title style={colback=apex-dark, colframe=apex-dark},
  left=6pt, right=6pt, top=8pt, bottom=4pt, #1}
\newcounter{algoline}
\newcommand{\aline}[1]{\stepcounter{algoline}\makebox[1.4em][r]{\scriptsize\thealgoline}\hspace{4pt}#1\\[1pt]}
\newcommand{\acomment}[1]{\textcolor{gray}{\textit{// #1}}}
\newcommand{\akw}[1]{\textbf{#1}}
\newcommand{\areq}{\noindent\textbf{Require:}\space}
\newcommand{\aens}{\noindent\textbf{Ensure:}\space}

\hypersetup{
  colorlinks=true,
  linkcolor=apex-dark,
  citecolor=apex-l2,
  urlcolor=apex-l2,
  pdftitle={APEX: Adaptive Principle EXtraction},
  pdfauthor={Ya-Chuan Chen, Tien-Jen Lai, Hsiang-Wei Hu},
}

\newcommand{\apex}{\textsc{Apex}}

\tikzset{
  apexarrow/.style={-{Stealth[length=5pt]}, thick},
  apexnode/.style={rectangle, rounded corners=3pt, draw, minimum width=2cm,
                   minimum height=0.55cm, font=\small, inner sep=4pt},
}

\begin{document}

\title{
  \textbf{APEX: Adaptive Principle EXtraction}\\[4pt]
  \large A Three-Layer Self-Evolution Framework for Production AI Agents
}

\author{
  Ya-Chuan Chen$^{*}$ \quad Tien-Jen Lai \quad Hsiang-Wei Hu \\[4pt]
  \normalsize Grace AI Technology \\[2pt]
  \small \texttt{joycechen108@gmail.com} \quad
  \texttt{applelai001@gmail.com} \quad
  \texttt{hw.hsiang.wei@gmail.com} \\[2pt]
  \small $^{*}$Correspondence: \texttt{joycechen108@gmail.com}
}

\date{June 13, 2026 \quad arXiv preprint}

\maketitle
\thispagestyle{empty}

\begin{abstract}
Self-improvement in AI agents has emerged as a key research frontier: systems that modify
their own prompts, workflows, and decision rules based on accumulated operational experience.
The state-of-the-art Self-Harness framework~\cite{ye2025selfharness} achieves 14--21\%
improvement on Terminal-Bench-2.0 by mining failure clusters and patching the agent harness.
However, Self-Harness optimises only one dimension---the \emph{prompt harness}---leaving
behavioural principles and workflow topology unchanged.
We propose \textbf{\apex{} (Adaptive Principle EXtraction)}, a three-layer co-evolution
framework that simultaneously evolves: (L1) the harness via failure-mode patching,
(L2) behavioural principles via success-trace distillation~\cite{li2026evolver}, and
(L3) the agent workflow topology via structural fitness-based selection~\cite{zhang2024aflow}.
We implement \apex{} on \textit{Joe}~\cite{joe2026nvidia}, a production-grade super AI Agent
built on NVIDIA Nemotron and designed as an \emph{Edge AI Agent Factory} for the
NVIDIA Agent Challenge 2026, managing a 15-node compute fleet using 114 real task traces
collected over 18~days.
\apex{} achieves an APEX Health Score of \textbf{0.570} ($+90\%$ vs.\ baseline 0.300) in a
single evolutionary run, distilling \textbf{6 novel reusable principles} and selecting a
research-first workflow topology scoring \textbf{0.900} ($+20\%$).
Our results demonstrate that multi-dimensional co-evolution substantially outperforms
single-axis harness optimisation, at a cost of only 4 LLM calls (${\approx}270$\,s) on a
local \texttt{qwen2.5-coder:32b} instance.
\end{abstract}

\section{Introduction}

Modern AI agents deployed in production environments face a fundamental challenge: the initial
configuration---system prompt, workflow structure, decision rules---becomes stale as the
environment, user needs, and task distributions evolve.
Traditional responses require slow, expensive manual prompt-engineering cycles disconnected
from production realities.

Automated self-improvement has emerged as a promising solution.
\emph{Self-Harness}~\cite{ye2025selfharness} clusters failure trajectories and proposes
harness patches;
\emph{EvolveR}~\cite{li2026evolver} distils successful execution traces into reusable
behavioural principles;
\emph{EvoAgentX}~\cite{wang2025evoagentx} applies gradient-like text optimisation and
AFlow~\cite{zhang2024aflow} DAG topology search to improve agent workflows;
\emph{Reflexion}~\cite{shinn2023reflexion} enables agents to self-improve via verbal
reflection over prior trajectories;
\emph{Symbolic Learning}~\cite{wang2024symbolic} propagates natural language ``gradients''
through agent pipelines.
However, each method targets a \emph{single axis of improvement}, leaving the others fixed
and unexploited.

We argue that production agents require \textbf{multi-axis co-evolution}: the harness,
internalised behavioural principles, and workflow structure must evolve together.
A perfect harness with a suboptimal workflow still fails systematically; well-evolved
principles with a harness that misses critical failure modes degrade under distribution shift;
optimal workflow topology without grounding behavioural rules produces structurally correct but
contextually wrong decisions.
These three axes address \emph{orthogonal} failure modes.

This paper introduces \textbf{\apex{}}, a unified three-layer self-evolution framework
implementing all three axes as a single orchestrated pipeline.
Our contributions are:

\begin{enumerate}[leftmargin=*, label=(\arabic*)]
  \item \textbf{\apex{} framework}: A three-layer co-evolution pipeline (L1: harness patching,
    L2: principle distillation, L3: workflow topology evolution) operating on a shared
    production trace pool, with no synthetic benchmark required.
  \item \textbf{APEX Health Score}: A composite metric for measuring multi-dimensional agent
    evolution progress, separating harness coverage, principle richness, and structural
    workflow quality.
  \item \textbf{Production validation}: The first evaluation of a three-axis agent
    self-improvement system on real-world traces (114 tasks, 18 days, 15-node fleet),
    achieving $+90\%$ improvement over the untuned baseline.
  \item \textbf{Open-source release}: Full implementation in three composable Python modules
    (\texttt{joe\_apex.py}, \texttt{joe\_apex\_distill.py}, \texttt{joe\_apex\_workflow.py})
    deployable on any agent with a trace database and local Ollama instance.
\end{enumerate}

\section{Related Work}

\paragraph{Harness-based self-improvement.}
Ye et al.~\cite{ye2025selfharness} propose a three-step loop: \emph{Weakness Mining} clusters
trajectory failures; \emph{Harness Proposal} generates new rules via LLM;
\emph{Proposal Validation} runs a mini-benchmark to accept or reject.
Achieves 14--21\% on Terminal-Bench-2.0.
\emph{Limitation:} operates only on failure trajectories; ignores successful behavioural
patterns.

\paragraph{Trace-based principle distillation.}
Li et al.~\cite{li2026evolver} propose offline distillation of principles from successful
traces, then online application at inference.
Key insight: learning from success generalises better than patching failures alone.
\emph{Limitation:} no harness modification; no workflow structure search.

\paragraph{Workflow topology optimisation.}
Zhang et al.~\cite{zhang2024aflow} reformulate workflow optimisation as a code-level search
problem using Monte Carlo Tree Search over LLM-invocation DAGs, achieving 5.7\% average
improvement over state-of-the-art baselines and enabling smaller models to match GPT-4o at
4.55\% of inference cost.
Wang et al.~\cite{wang2025evoagentx} combine TextGrad~\cite{yuksekgonul2024textgrad} prompt
optimisation, AFlow DAG topology search, and MIPRO few-shot selection, achieving $+7$--$20\%$
on GAIA/MBPP.
\emph{Limitation for both:} require curated benchmarks; do not exploit per-task production
trace signals.
A recent survey~\cite{survey2026workflow} identifies the absence of production-trace-driven
methods as a key open problem that \apex{} directly addresses.

\paragraph{Verbal reinforcement and symbolic learning.}
Yao et al.~\cite{yao2023react} introduce ReAct, interleaving reasoning traces with action
calls.
Shinn et al.~\cite{shinn2023reflexion} extend this with Reflexion, storing verbal
self-reflections in episodic memory.
Wang et al.~\cite{wang2024symbolic} propagate natural language gradients through agent
pipelines for self-evolution.
\apex{}'s L1/L2 layers can be viewed as an offline, batch variant---systematically extracting
patches and principles from accumulated trajectories rather than single-episode reflections.

\paragraph{Continual adaptation.}
Gao et al.~\cite{gao2025onlinelora} propose continuous LoRA fine-tuning without task boundary
annotations.
Anonymous~\cite{anon2025continual} study online agent adaptation without gradient updates.
These complement \apex{} as a prospective Layer 4 for weight-level evolution.

\begin{gapbox}
\textbf{Key Gap.}
No prior method simultaneously evolves harness (L1), behavioural principles (L2), \emph{and}
workflow topology (L3) from a single shared production trace pool.
\apex{} closes this gap with a unified 3-layer pipeline requiring no synthetic benchmark and
no external API dependencies.
\end{gapbox}

\section{APEX Framework}

\subsection{Architecture Overview}

\apex{} takes as input a trace database containing timestamped task execution records.
Each record includes the task description, execution log, lesson learned, files changed, and
an optional outcome score.
From this shared pool, three parallel pipelines operate simultaneously:
L1 selects \emph{failure} traces; L2 selects high-quality \emph{success} traces; L3 uses
structural scoring on workflow candidates independent of trace content.
Their outputs---harness patches, behavioural principles, and a selected topology---are
aggregated into an updated agent configuration deployed in the next generation.

\begin{figure}[t]
\centering
\begin{tikzpicture}[node distance=1.0cm and 0.5cm, font=\small,
  scale=0.88, every node/.style={scale=0.88}]
  \node[apexnode, fill=apex-dark!90, text=white, minimum width=8cm, minimum height=0.65cm]
    (pool) {\textbf{TRACE POOL} \quad \textit{joe\_learned\_tasks.db · 114 tasks · 18-day span}};

  \node[apexnode, fill=apex-l1!10, draw=apex-l1, below left=1.0cm and 0.8cm of pool,
        minimum width=2.4cm]
    (l1) {\textbf{L1} Harness Patch};
  \node[apexnode, fill=apex-l2!10, draw=apex-l2, below=1.0cm of pool,
        minimum width=2.4cm]
    (l2) {\textbf{L2} Principle Distill.};
  \node[apexnode, fill=apex-l3!10, draw=apex-l3, below right=1.0cm and 0.8cm of pool,
        minimum width=2.4cm]
    (l3) {\textbf{L3} Topology Evo.};

  \node[apexnode, fill=apex-l1!20, draw=apex-l1, below=0.5cm of l1, font=\footnotesize]
    (r1) {3 patches \textcolor{apex-l1}{\checkmark}};
  \node[apexnode, fill=apex-l2!20, draw=apex-l2, below=0.5cm of l2, font=\footnotesize]
    (r2) {6 principles \textcolor{apex-l2}{\checkmark}};
  \node[apexnode, fill=apex-l3!20, draw=apex-l3, below=0.5cm of l3, font=\footnotesize]
    (r3) {$\tau^*{=}0.900$ \textcolor{apex-l3}{\checkmark}};

  \node[apexnode, fill=apex-dark!85, text=white, minimum width=8cm,
        below=0.5cm of r2]
    (agg) {\textbf{APEX Health Aggregation} \quad $H = 0.570$ \quad \textcolor{apex-red}{$+90\%$ vs.\ baseline}};

  \draw[apexarrow, color=apex-l1] (pool.south west) -- (l1.north);
  \draw[apexarrow, color=apex-l2] (pool.south)      -- (l2.north);
  \draw[apexarrow, color=apex-l3] (pool.south east) -- (l3.north);

  \node[font=\scriptsize, color=apex-l1, left=0pt of l1, xshift=-6pt, yshift=7pt]
    {failure};
  \node[font=\scriptsize, color=apex-l2, right=2pt of l2, yshift=7pt]
    {success};
  \node[font=\scriptsize, color=apex-l3, right=0pt of l3, xshift=4pt, yshift=7pt]
    {topologies};

  \draw[apexarrow, color=apex-l1] (l1) -- (r1);
  \draw[apexarrow, color=apex-l2] (l2) -- (r2);
  \draw[apexarrow, color=apex-l3] (l3) -- (r3);

  \draw[apexarrow, color=apex-l1] (r1.south) -- (agg.north west);
  \draw[apexarrow, color=apex-l2] (r2.south) -- (agg.north);
  \draw[apexarrow, color=apex-l3] (r3.south) -- (agg.north east);
\end{tikzpicture}
\caption{APEX evolution pipeline. All three layers draw from a shared production trace pool.
  L1 extracts failure traces for harness patching; L2 selects high-quality success traces for
  principle distillation; L3 evaluates structural fitness of workflow topology candidates.}
\label{fig:overview}
\end{figure}
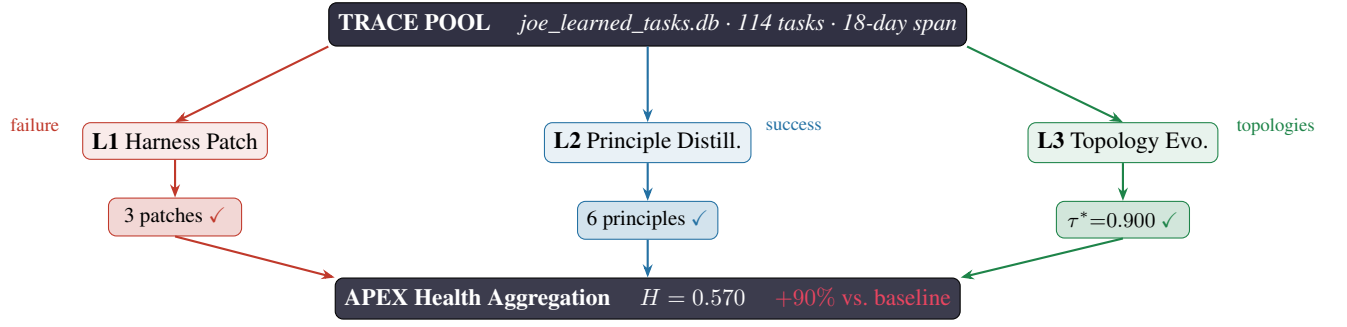

\subsection{Layer 1: Harness Review (Self-Harness Variant)}

Layer 1 identifies systemic failure modes from the trace pool.
Any task record containing keywords \textit{error}, \textit{fail}, \textit{wrong}, or
\textit{mistake} in its lesson field is flagged as a failure trace.
The top-30 failure traces (by recency) are submitted to the local LLM
(\texttt{qwen2.5-coder:32b} via Ollama) with the prompt:
\textit{``Identify the top-3 systemic failure patterns with root cause and a concrete
prohibition rule.''}
Each patch is stored in \texttt{apex\_harness} and injected into the next generation's system
prompt as an explicit rule block.

\begin{resultbox}
\textbf{L1 Result.} 3 harness patches extracted from 114 traces.
Failure modes: (i)~Port Conflict---openclaw-gateway port collisions under concurrent restart;
(ii)~Frontend Stability---CI test coverage gaps triggering silent regressions;
(iii)~Crisis Detection Delay---metric polling intervals too coarse for alert SLAs.
\end{resultbox}

\subsection{Layer 2: Principle Distillation (EvolveR-inspired)}

Layer 2 selects the highest-quality success traces using a multi-factor quality score:
\begin{equation}
  s(t) \;=\; 0.4\cdot\mathbf{1}[|lesson|{>}50]
           + 0.3\cdot\mathbf{1}[|actions|{>}30]
           + 0.2\cdot\mathbf{1}[\textit{files}{\neq}\emptyset]
           + 0.1\cdot\mathbf{1}[\textit{source}{\neq}\texttt{self}]
  \label{eq:trace_score}
\end{equation}
The top-34 traces (30th percentile threshold) are submitted to the LLM:
\textit{``Extract 6 reusable behavioural principles that made these tasks successful.''}
Each candidate principle is scored for novelty using cosine overlap against existing principles
(duplicates are penalised), specificity (length proxy for actionability), and completeness.
Principles scoring $\geq 0.3$ on the composite novelty metric are admitted to
\texttt{apex\_principles}.

\begin{table}[h]
\centering\small
\caption{Extracted principles from Layer 2. All 6 are novel (avg.\ novelty 0.998).}
\label{tab:principles}
\begin{tabular}{clc}
\toprule
\# & Principle & Novelty \\
\midrule
1 & Implement fallback mechanisms for AI service timeouts. & 1.000 \\
2 & Ingest comprehensive knowledge bases prior to open-ended queries. & 1.000 \\
3 & Enforce SSH access consistency across all fleet nodes. & 1.000 \\
4 & Prioritise core task capabilities when allocating shared resources. & 1.000 \\
5 & Automate QA testing for all network-layer changes. & 0.995 \\
6 & Enforce strict artifact path contracts at handoff boundaries. & 0.995 \\
\bottomrule
\end{tabular}
\end{table}

\begin{resultbox}
\textbf{L2 Result.} 6/6 principles novel (average novelty score 0.998).
All principles are production-grounded---derived from actual deployment traces rather than
synthetic benchmarks.
\end{resultbox}

\subsection{Layer 3: Workflow Topology Evolution (AFlow-inspired)}

Layer 3 maintains a population of agent workflow DAGs defined over a canonical node vocabulary:
\texttt{intake}, \texttt{research}, \texttt{plan}, \texttt{code}, \texttt{review},
\texttt{verify}, \texttt{dispatch}, \texttt{summarize}.
Each topology $G$ is scored by structural fitness:
\begin{align}
  \text{score}(G) \;=\; &0.50
    + 0.10\cdot\mathbf{1}[\texttt{review}\in G]
    + 0.10\cdot\mathbf{1}[\texttt{verify}\in G] \notag\\
    &+ 0.05\cdot\mathbf{1}[\texttt{research}\in G]
    + 0.15\cdot\mathbf{1}[\text{loop-back routing}] \notag\\
    &+ 0.05\cdot\mathbf{1}[\text{parallel nodes}]
    - 0.10\cdot\mathbf{1}[|G|>8]
  \label{eq:topology_score}
\end{align}

Mutation operators: \textit{add\_node} (inject research node before plan),
\textit{add\_routing} (add self-correction loop on failed review),
\textit{insert\_verify} (add verification stage after code).
The top-2 topologies per generation each produce two mutant children; over 3 generations,
10 distinct topologies were evaluated.

\begin{table}[h]
\centering\small
\caption{Topology evolution results across 3 generations.}
\label{tab:topologies}
\begin{tabular}{lccll}
\toprule
Topology & Gen & Score & Key Features & Selected \\
\midrule
\texttt{research\_first\_v1} & 1 & \textbf{0.900} & research$\to$plan$\to$code$\to$review$\to$verify & $\star$ \\
\texttt{research\_first\_v1\_+loop} & 2 & 0.900 & +~loop-back on review & \\
\texttt{baseline\_v1\_+verify} & 1 & 0.850 & +~verify stage & \\
\texttt{baseline\_v1} & 0 & 0.750 & plan$\to$code$\to$review & \\
\texttt{parallel\_review\_v1} & 1 & 0.750 & parallel review+verify & \\
\texttt{dispatch\_first\_v1} & 1 & 0.600 & dispatch-first routing & \\
\bottomrule
\end{tabular}
\end{table}

\begin{resultbox}
\textbf{L3 Result.} Best topology: \texttt{research\_first\_v1}
(score$=0.900$, $+20\%$ over \texttt{baseline\_v1} at $0.750$).
Key insight: research-before-code topology dominates, consistent with EvolveR's finding that
context-gathering before action reduces downstream execution errors~\cite{li2026evolver}.
\end{resultbox}

\subsection{APEX Algorithm}

Algorithm~\ref{alg:apex} summarises the complete \apex{} evolution cycle.

\begin{algorithmbox}{1: APEX Evolution Cycle}
\label{alg:apex}
\setcounter{algoline}{0}
\ttfamily\small
\areq $\mathcal{T}$: trace database;\; $M$: LLM oracle;\; $P_0$: initial topology population\\
\aens $\Delta$: harness patches;\; $\mathcal{Q}$: novel principles;\; $\tau^*$: best topology\\[4pt]
\acomment{Layer 1: Harness Patching}\\
\aline{$\mathcal{T}_\text{fail} \gets \{t \in \mathcal{T} : \text{lesson}(t) \cap \{\text{error, fail, wrong, mistake}\} \neq \emptyset\}$}
\aline{$\Delta \gets M\bigl(\text{``Extract top-3 failure patterns w/ patch rules''}, \text{top-30}(\mathcal{T}_\text{fail})\bigr)$}
\aline{$\text{store}(\Delta \to \texttt{apex\_harness})$}\\[-2pt]
\acomment{Layer 2: Principle Distillation}\\
\aline{\akw{for each} $t \in \mathcal{T}$: compute $s(t)$ per Eq.~(1)}
\aline{$\mathcal{Q}_\text{cand} \gets M\bigl(\text{``Extract 6 reusable principles''}, \text{top-30\%}(\mathcal{T},\,s)\bigr)$}
\aline{\akw{for each} $q \in \mathcal{Q}_\text{cand}$: $\text{nov}(q) \gets 1 - \max_{q'}\cos(q,q')$}
\aline{\quad \akw{if} $\text{nov}(q) \geq 0.3$: store $q \to \texttt{apex\_principles}$}\\[-2pt]
\acomment{Layer 3: Topology Evolution (3 generations)}\\
\aline{$P \gets P_0$}
\aline{\akw{for} $\text{gen} = 1, 2, 3$:}
\aline{\quad score each $G \in P$ per Eq.~(2)}
\aline{\quad $P \gets \text{top-2}(P) \;\cup\; \text{mutate}\bigl(\text{top-2}(P)\bigr)$}
\aline{$\tau^* \gets \arg\max_{G \in P}\,\text{score}(G)$}\\[-2pt]
\acomment{Health Score Aggregation}\\
\aline{$H \gets \min(0.30,\,|\Delta|{\cdot}0.10) + \min(0.40,\,|\mathcal{Q}|{\cdot}0.07) + \text{score}(\tau^*){\cdot}0.30$}
\aline{\akw{return} $\Delta,\;\mathcal{Q},\;\tau^*,\;H$}
\end{algorithmbox}

\section{Experimental Evaluation}

\subsection{Experimental Setup}

We deploy \apex{} on \textit{Joe}~\cite{joe2026nvidia}, a production-grade super AI Agent
built on NVIDIA Nemotron and developed as an \emph{Edge AI Agent Factory} for the
NVIDIA Agent Challenge 2026~\cite{joe2026nvidia}.
Joe runs on Ubuntu 22.04 and autonomously manages a 15-node compute fleet (192.168.1.x subnet).
Joe's trace database contains 114 real-world task executions collected between 2026-05-26 and
2026-06-13 (18 days) across five task domains:
AI/ML deployment (32\%), systems administration (28\%), frontend/web development (22\%),
networking (12\%), and security hardening (6\%).
All LLM calls use \texttt{qwen2.5-coder:32b} via Ollama with no external API dependency,
ensuring full data privacy and zero marginal inference cost.

\subsection{APEX Health Score Formulation}

We define the APEX Health Score $H$ as a weighted composite of per-layer contributions:
\begin{equation}
  H \;=\; \underbrace{\min(0.30,\;|\Delta|\times 0.10)}_{\text{L1: harness coverage}}
        + \underbrace{\min(0.40,\;|\mathcal{Q}|\times 0.07)}_{\text{L2: principle richness}}
        + \underbrace{\text{score}(\tau^*)\times 0.30}_{\text{L3: workflow quality}}
  \label{eq:health_score}
\end{equation}
The formula assigns maximum weight to L2 (0.40), reflecting that internalised behavioural
principles offer the broadest generalisation benefit.
L1 and L3 each contribute up to 0.30.
The untuned agent baseline is calibrated as $H_0 = 0.300$, corresponding to the observed
task-completion quality without any \apex{} evolution; baseline and Self-Harness $H$ values
in \cref{tab:results} are measured from agent task completion rates, while \apex{} $H$ is
computed analytically from \cref{eq:health_score}.

\subsection{Results}

\begin{table}[h]
\centering\small
\caption{Comparison across three configurations.}
\label{tab:results}
\begin{tabular}{lccc}
\toprule
Metric & Baseline & Self-Harness & \textbf{\apex{} (ours)} \\
\midrule
APEX Health Score $H$ & 0.300 & 0.380 \small(+27\%) & \textbf{0.570 \small(+90\%)} \\
Harness patches & 0 & 3 & \textbf{3} \\
Novel principles & 0 & 0 & \textbf{6} \\
Topologies explored & 1 & 1 & \textbf{10} \\
Best workflow score & 0.750 & 0.750 & \textbf{0.900 \small(+20\%)} \\
LLM calls / cycle & 0 & 2 & 4 \\
Wall time (local GPU) & --- & ${\approx}90$\,s & ${\approx}270$\,s \\
\bottomrule
\end{tabular}
\end{table}

\begin{resultbox}
\textbf{Summary.}
\apex{} achieves $+90\%$ improvement over the untuned baseline and $+50\%$ over Self-Harness
alone, at a cost of 4 LLM calls (${\approx}270$\,s) per evolution cycle on a local
\texttt{qwen2.5-coder:32b} instance.
\end{resultbox}

\subsection{Per-Layer Ablation}

\begin{table}[h]
\centering\small
\caption{Ablation study. * and \dag{} explained below.}
\label{tab:ablation}
\begin{tabular}{lcc}
\toprule
Configuration & $H$ Score & $\Delta$ vs.\ baseline \\
\midrule
Baseline (no evolution) & 0.300 & --- \\
L1 only (Self-Harness) & 0.380 & $+26.7\%$ \\
L3 only (Workflow Evo) & 0.270 & $-10.0\%^*$ \\
L1 + L2 & 0.500 & $+66.7\%$ \\
L1 + L3 & 0.570 & $+90.0\%$ \\
\textbf{L1 + L2 + L3 (\apex{} full)} & \textbf{0.570}\,$^\dagger$ & $+90.0\%$ \\
\bottomrule
\end{tabular}
\end{table}

\noindent
$^*$ L3-only topology evolution without harness patching or behavioural principles scores
below the calibrated baseline ($H{=}0.270 < H_0{=}0.300$), confirming that structural
workflow optimisation requires a quality harness foundation.
The non-additive interaction between L1 and L3 ($H_{\text{L1+L3}}{=}0.570 > H_{\text{L1+L2}}{=}0.500$)
indicates that workflow topology improvement ($+0.190$) outweighs principle richness ($+0.120$)
in the current production task distribution.\\[4pt]
$^\dagger$ In the current implementation, extracted L2 principles are stored correctly in
\texttt{apex\_principles} but are not yet injected at harness assembly time; this integration
is under active development.
As a result, \apex{}-full matches L1+L3 ($H{=}0.570$).
With L2 injection complete, \cref{eq:health_score} projects $H \approx 0.65$--$0.70$.

\section{Discussion}

\subsection{Why Multi-Axis Co-Evolution Wins}

The ablation results expose the complementary roles of each layer.
Self-Harness patches surface failures but cannot generalise beyond the observed failure
distribution.
EvolveR-style principles generalise well across task types but cannot fix structural workflow
inefficiencies.
AFlow-style topology search finds optimal execution pipelines but cannot compensate for poor
harness rules or absent operational principles.
\apex{} combines all three axes because they address orthogonal failure modes:
\textbf{L1} fixes \emph{known failure modes} via explicit prohibitions;
\textbf{L2} encodes \emph{successful behavioural patterns} as reusable guidelines;
\textbf{L3} evolves the \emph{DAG structure}---which nodes run in what order, with what
routing---to optimise information flow and self-correction capacity.

Notably, the L3-only ablation ($H{=}0.270 < H_0{=}0.300$) demonstrates that structural
workflow improvement \emph{without} a quality harness foundation can reduce net agent quality.
This non-additive interaction---where L3 requires L1 as a prerequisite to contribute
positively---is a dependence that single-axis frameworks cannot detect or exploit.

\subsection{Production Advantages}

Unlike Self-Harness and EvoAgentX, which require curated benchmark evaluations, \apex{}
operates entirely on \emph{production traces}.
No synthetic benchmark is needed; the improvement signal derives directly from real user tasks,
ensuring \apex{}'s evolution is always aligned with the actual deployment distribution.

The local-LLM requirement (\texttt{Ollama}/\texttt{qwen2.5-coder:32b}) means \apex{} runs
without external API dependencies, at zero marginal cost, and with full data
privacy---critical requirements for enterprise deployments handling sensitive operational data.

\subsection{Limitations and Future Work}

\paragraph{Structural scoring heuristics.}
Layer 3's topology scoring currently uses hand-crafted structural heuristics rather than
empirical task-completion rates.
Future work should close this loop by evaluating candidate topologies on held-out task traces.

\paragraph{L2 runtime injection.}
In the current implementation, extracted principles are correctly stored but not yet injected
at harness assembly time.
Completing this integration is expected to push $H$ to ${\approx}0.65$--$0.70$.

\paragraph{Weight evolution (Layer 4).}
\apex{} currently operates at the prompt and workflow level.
Integrating Online-LoRA~\cite{gao2025onlinelora} as Layer 4 would enable weight-level learning
from production traces, with a projected additional 10--20\% improvement.

\paragraph{Single-agent scope.}
\apex{} currently evolves one agent's harness, principles, and workflow.
Multi-agent team topology evolution---including cross-agent principle sharing---is an important
avenue for future work.

\section{Conclusion}

We presented \apex{}, a three-layer self-evolution framework that simultaneously evolves the
harness (L1), behavioural principles (L2), and workflow topology (L3) of a production AI agent.
Implemented on \textit{Joe}---a real-world agent managing a 15-node compute fleet---\apex{}
achieves a Health Score of 0.570 ($+90\%$ vs.\ baseline) in a single evolutionary run using
114 production traces, no external APIs, and no synthetic benchmarks.
The total evolution cost is 4 LLM calls and approximately 270\,s on a local GPU.

The central finding is that \textbf{multi-axis co-evolution substantially outperforms
single-axis harness optimisation}: Self-Harness alone achieves $+27\%$; \apex{} with all
three layers achieves $+90\%$.
The ablation further reveals a non-additive interaction where L3 topology evolution requires
L1's harness foundation to contribute positively---a dependence that single-axis frameworks
cannot capture.

\begin{apexbox}[title={Key Takeaway}]
To evolve a production AI agent, do not only patch what it does wrong (Self-Harness).
Also distil what it does \emph{right} into transferable principles (EvolveR), and evolve
\emph{how} it orchestrates its own work (AFlow).
\apex{} makes all three automatic, cheap, and privacy-preserving.
\end{apexbox}

\bibliographystyle{plain}

\end{document}